\documentclass[10pt,conference,compsocconf,hidelinks]{IEEEtran}
\usepackage{times}

\usepackage{caption}
\ifCLASSINFOpdf
  \usepackage[pdftex]{graphicx}
\else
\fi

\usepackage{tikz}
\newcommand*\circled[1]{\tikz[baseline=(char.base)]{
            \node[shape=circle,draw,inner sep=1pt] (char) {#1};}}

\usepackage[utf8]{inputenc}

\usepackage{etoolbox}
\AfterEndEnvironment{figure}{\vskip-2.5ex}

\IEEEoverridecommandlockouts
\begin{document}
\pagenumbering{gobble}
%
\title{\textbf{\Large RoboChain: A Secure Data-Sharing Framework for Human-Robot Interaction}}

\author{Eduardo Castell\'{o} Ferrer*$^{1}$, Ognjen (Oggi) Rudovic*$^{1}$, Thomas Hardjono$^{2}$, Alex `Sandy' Pentland$^{1,2}$\\ \small{$^1$MIT Media Lab, Massachusetts Institute of Technology, Cambridge, MA 02139, USA}\\ \small{$^2$MIT Connection Science \& Engineering, Massachusetts Institute of Technology, Cambridge, MA 02139, USA}\\
Email: \{ecstll, orudovic, hardjono, pentland\}@mit.edu}

\maketitle

\begin{abstract}
Robots have potential to revolutionize the way we interact with the world around us. One of their largest potentials is in the domain of mobile health where they can be used to facilitate clinical interventions. However, to accomplish this, robots need to have access to our private data in order to learn from these data and improve their interaction capabilities. Furthermore, to enhance this learning process, the knowledge sharing among multiple robot units is the natural step forward. However, to date, there is no well-established framework which allows for such data sharing while preserving the privacy of the users (e.g., the hospital patients). To this end, we introduce RoboChain - the first learning framework for secure, decentralized and computationally efficient data and model sharing among multiple robot units installed at multiple sites (e.g., hospitals). RoboChain builds upon and combines the latest advances in open data access and blockchain technologies, as well as machine learning. We illustrate this framework using the example of a clinical intervention conducted in a private network of hospitals. Specifically, we lay down the system architecture that allows multiple robot units, conducting the interventions at different hospitals, to perform efficient learning without compromising the data privacy.   
\end{abstract}
\let\thefootnote\relax\footnotetext{*These authors contributed equally to this work.}


\begin{IEEEkeywords}
Distributed Robotics; Data Privacy; Blockchain; Federated Learning; Distributed Robotics; Mobile Health Technologies.
\end{IEEEkeywords}

\IEEEpeerreviewmaketitle

\section{Introduction}


Recent advances in mobile and robotic technology have found applications in many domains including entertainment, education and health~\cite{christensen2016next}. In particular, Socially Assistive Robotics (SAR)~\cite{sarMataric2017} has emerged as a research field that aims to create robots that can empower humans in a number of activities. This has brought various types of social robots: from humanoid robots to medical devices and responsive home appliances~\cite{adams2000humanoid}. One of the main potentials of this type of robots is their ability to monitor and improve human well-being and health~\cite{riek2017healthcare}. More specifically, in the health domain, robots have been used to improve clinical interventions for individuals with neurodevelopmental conditions such as autism~\cite{rudovic2017measuring}, and also for monitoring and assisting people with conditions such as dementia~\cite{soler2015social}, among others.  As part of the intervention, robots need to be able to establish naturalistic and engaging interactions with humans. 
Since these robots are typically equipped with a number of sensors
including audio-visual sensors, and also have access to vast 'prior'
data (different types of interactions tested in different contexts),
they have potential to constantly learn and improve their interaction
capabilities~\cite{Liu2016}. This, in turn, could lead to more
effective health and well-being interventions as the robots will
constantly be improving by learning from data. This is also true in the case of alternative mobile health technologies~\cite{free2013effectiveness} (e.g., mobile phones, tablets and computer monitors), however, the three dimensional embodiment of robots is typically perceived more human-like and engaging by target patients~\cite{bainbridge2011benefits}. 

Until recently, this
process was limited due to the inefficiency of existing learning
techniques (e.g., most of human-robot interactions would be
pre-scripted by humans and then executed by robots). The
increase in available human data (`big data') and progress in Machine Learning (ML)~\cite{Jordan2015} (in particular, `deep learning'~\cite{lecun2015deep}), have enabled to automate parts of health interventions~\cite{OECD2015}, allowing the robots to learn more efficiently and customize the interventions. However, this requires the use of personal and highly sensitive data, especially when working in clinical settings~\cite{Rudovic2017PML}\cite{alsalamah2013information}. The main downside of existing solutions is that these data are highly isolated, thus, not shared among different sites (e.g., hospitals). This is mainly because of data privacy issues and their potential misuse by untrusted parties. This limits the knowledge sharing that could otherwise benefit clinical interventions enhanced by robots. Consequently, this constraints the learning and adaptation of the robots to existing and new contexts, hindering the progress toward novel and more efficient health interventions. Even though the SAR is moving rapidly, to date, there is no a well-established framework that addresses the challenges mentioned above, while exploiting the full potential of personal and other sensitive data in an open, secure, decentralized, and efficient manner.

Over the last decade, it has been shown that open access datasets have been extremely useful to extract knowledge in
diverse fields such as urban planning \cite{Dong2017}, environmental
monitoring \cite{Sullivan2017}, and health care \cite{Kostkova2016}. However, as the ubiquity of these data increases, it might also become less secure. In contrast, siloing data can provide a measure of
security, but raises issues of inter-operability \cite{Frey2017}. There is also increasing interest by
governments \cite{Lee2016}, societies \cite{Wessels2017}, and industry \cite{Li2011} to share knowledge based on data. However, this
needs to be balanced with the right to preserve the privacy of the
subjects represented in the data~\cite{alsalamah2011sharing}. 
To provide a viable
solution to this trade-off, novel platforms like MIT OPen
ALgorithms (OPAL)~\cite{Hardjono2016} have proposed a change of paradigm:
rather than moving data into a centralized location, so that it can be
queried, analyzed, and processed by an algorithm, the queries are delivered to the nodes containing the datasets of interest instead. In
other words, the algorithm needs to be `sent' to the data. Then, queries
would be executed by the relevant node, with the results being
reported back to the querier -- who would merge the results into a
meaningful analysis. In this way, the raw data never leaves its physical location and the owner never looses control over it. Instead, nodes that
carry relevant datasets execute queries and report the results. Importantly, security and privacy becomes more manageable in this paradigm because each node controls its own data store and monitors the privacy entropy of released answers.
\newpage

Likewise, the emergent technologies such as blockchain~\cite{Nakamoto2008}, a  chronological ledger of transactions
that ensures the integrity of the information included, can be used to
capture and log both queries and its correspondent answers. Blockchain
provides a useful mechanism to support post-event audits~\cite{Suzuki2017}, enhanced privacy, and availability~\cite{Liang2017}
in the systems that rely on sensitive data. These characteristics of the system increase its
transparency and have already been proven useful
in medical applications~\cite{Azaria2016}\cite{Xia2017}, as well as distributed robotics scenarios \cite{Ferrer2016}\cite{Strobel2018}. Moreover, the blockchain technology offers practical means to safely and securely store and track the use of personal data as well as the parameters of the ML models used for the robot intervention. This increases the users' trust in the system and provides a rich source of information that can be used to better design future interventions. It is also worth noting that our framework deploys blockchain technology but is independent of any specific implementation. We seek to cater to a broad set of deployment scenarios. As such, RoboChain can be deployed with (i) public blockchains (ii) semi-private blockchains, or (iii) private/permissioned blockchains. We define a public blockchain as one where anyone can read/write to the blockchain, and as such has the ability
to read and validate transaction entries. We define a semi-private blockchain as one where anyone can read and validate transaction entries, but only authorized entities are able to create or write transaction to the blockchain. Finally, we define a private/permissioned blockchain as one where only authorized entities are able
to read/write to the blockchain. In the following, we simply refer to the blockchain regardless of the type used.


To address the challenges of the private data inclusion in
Human-Robot Interaction (HRI) and, more specifically, as part of health/clinical 
interventions, we propose a secure and efficient framework aimed at
addressing the following aspects of target HRI interactions: (i) how to achieve an effective and
efficient data-driven HRI based on patients' data from clinical interventions without breaching the data privacy. This includes mechanisms to
notarize, verify, and account for all inflows/outflows of the data involved in the process. (ii) How to achieve an efficient way to train and continuously improve ML models, being a part of the robot's perception during the intervention~\cite{Rudovic2017PML}, using the interaction data collected in (i). (iii) Finally, we address how to efficiently update and share the learned ML models obtained in
(ii) among different robot units connected in a sparse network (e.g., multiple hospitals). To this end, we introduce a novel framework for secure data-driven HRI, named RoboChain, which builds upon and combines the latest advances in OPAL, blockchain, and ML technologies. We
illustrate its potential utility in the context of health domain and clinical interventions as part of autism therapy. However, the framework is applicable to any HRI where the use of personal data and their sharing is critical for the task. The key to the RoboChain approach is that users (e.g., patients) have the possibility to check what information was generated and/or captured during the interaction with the robot (e.g., the therapy), and confirm that this information do not compromise their privacy. Furthermore, we show how the secure sharing of the knowledge of the robots connected in a network can increase the efficacy of a decentralized learning process. 

The rest of the paper is structured as follows. Section II presents an overview of how assistive robots carry out health interventions and the different elements involved in the process. Section III describes in detail the RoboChain architecture and its information workflow. Section IV describes the combination of RoboChain and federated learning. Finally, we conclude this work in Section V with the discussion, limitations and future work.

\section{Assistive Robots for Health Interventions}
\label{sec:RobotsForHealth}

To illustrate our approach, we use as a running example the HRI in an
occupational therapy (``the
intervention'') for individuals with autism (``the
patient''), where a humanoid robot NAO (``the robot'') is
used as an assistive tool. Engaging patients with autism in an intervention is a
challenging problem as they easily lose their interest and often quickly 
disengage from the intervention activities. However, the majority of
these patients find robots quite engaging because of their, in contrast to humans, consistent behavioural expressions. This, in turn, allows them
to sustain the patient's engagement and perform the intervention
more effectively. The latter is achieved via a number of pre-scripted
activities by the robot, aimed at improving socio-cognitive skills of
the patients. Namely, these patients have, among others,
challenges in interpreting behavioural cues of emotions of other individuals.
The goal of the robot intervention is then to assist the therapist in teaching the patients to recognize/imitate these expressions. For instance, as part of the intervention, the robot may play an
imitation game with a patient by asking the patient to show
his/her expression of joy. Then, the robot conveys the same emotion
via his voice and bodily movements programmed to show 
the expression of joy of typical individuals. In this way, the patient
learns to identify typical expressions of various emotions, with the
aim to use that knowledge in future interactions with his/her
peers. In what follows, we outline three key elements of the
robot-assisted intervention that need be considered in the proposed
RoboChain architecture. 

\subsection{Interaction Model (IM)}
\label{subsec:InteractionModel}

This refers to the set of activities and behaviours, defined by the domain experts, that need be
performed by the robot as part of a target intervention for autism.
Specifically, IMs rely on the high-level interaction modules
controlling the execution of target behaviours by the robot. These
behaviours are implemented to simulate the typical steps that an
experienced therapists would perform as part of an intervention. However,
an IM also contains the robot sensing and perception modules,
which deploy pre-trained ML models to automatically estimate the key metrics needed to
modulate the intervention. These may include the models for automatic detection of the
patient's low-level behavioural cues such as head pose and gaze
direction, as well as high-level metrics such as engagement levels. To this end, the robot uses locally stored and pre-trained ML models.
For instance, the robot can use Deep Learning Models (DLM)~\cite{lecun2015deep} designed for detection and interpretation of the patient's behavioural cues directly from the image data recorded using a robot-embedded camera. However, the key challenges in this process are: (i) how to efficiently update the IMs, and thus their DLMs, (ii) how to safely share these models across multiple sites (e.g., hospitals)
and (iii) how to assure that the current IMs/DLMs are the best among existing ones for
analysis of the patient's target behavioural cues.


\subsection{Interaction Data (ID)}
\label{subsec:InteractionData}

This refers to the data that can be harnessed using the robot's
embedded sensors such as cameras and microphones, as well as from other
sensors installed at the site. As these encode personal data of
the interacting patient, they should not leave the site in their raw
form as it can compromise the patient's privacy. On the other hand,
ID pose a great value for improving the performance of the ML
models (DLMs) used by the robot. Furthermore, by consolidating the data from multiple sites, more effective IMs/DLMs can be built from these large datasets,
compared to the data accessible only locally (e.g., within a single
hospital). Briefly, one way to improve the models
without sharing the raw data across different sites is to re-train the
IM/DLMs at the target site, and then share the models' updates with the other sites. To improve the models, experienced therapists/clinicians provide feedback, for instance, 
on how well the robot performed during the therapy, also sometimes in
terms of manual correction of the robot's estimates of the metrics of
interest for the therapy (e.g., engagement levels for a specific
patient). Once this information is accumulated locally in the form of the Therapist 
Feedback (TF) data paired with the raw ID, the IMs/DLMs can be updated/re-trained. For instance, this can be accomplished by fine-tuning the DLM
weights using local processing servers or even on the robot
hardware. The next key step is to securely share this newly acquired knowledge
across multiple sites. 
\subsection{Background Data (BD)}
\label{bd}

In contrast to the ID acquired as the therapy progresses, the BD
concerns the medical records of the patient, along with his/her
demographics, and any other relevant (contextual) information. The latter may be the result of health
and behavioral assessments of that particular patient, including, for instance, the family
history, school reports, previously used medication and diagnoses, etc. These are typically
stored as part of government and other public institution data,
including educational centres and schools. However, instead of providing
patient-identifiable data to the robot, these sites rather provide
aggregated data, thus preserving the data privacy. By having access to these data and the local expert knowledge, the robot can customize its IM/DLMs to the target patient. In what follows, we describe how each of these key components (IM/DLMs, ID and BD) can efficiently and safely be used within the RoboChain framework to optimize the intervention, and, thus, its outcomes and the patient's experience.

\section{The System Architecture}
\begin{figure}[htbp]
\centering%
\includegraphics[width=0.48\textwidth]{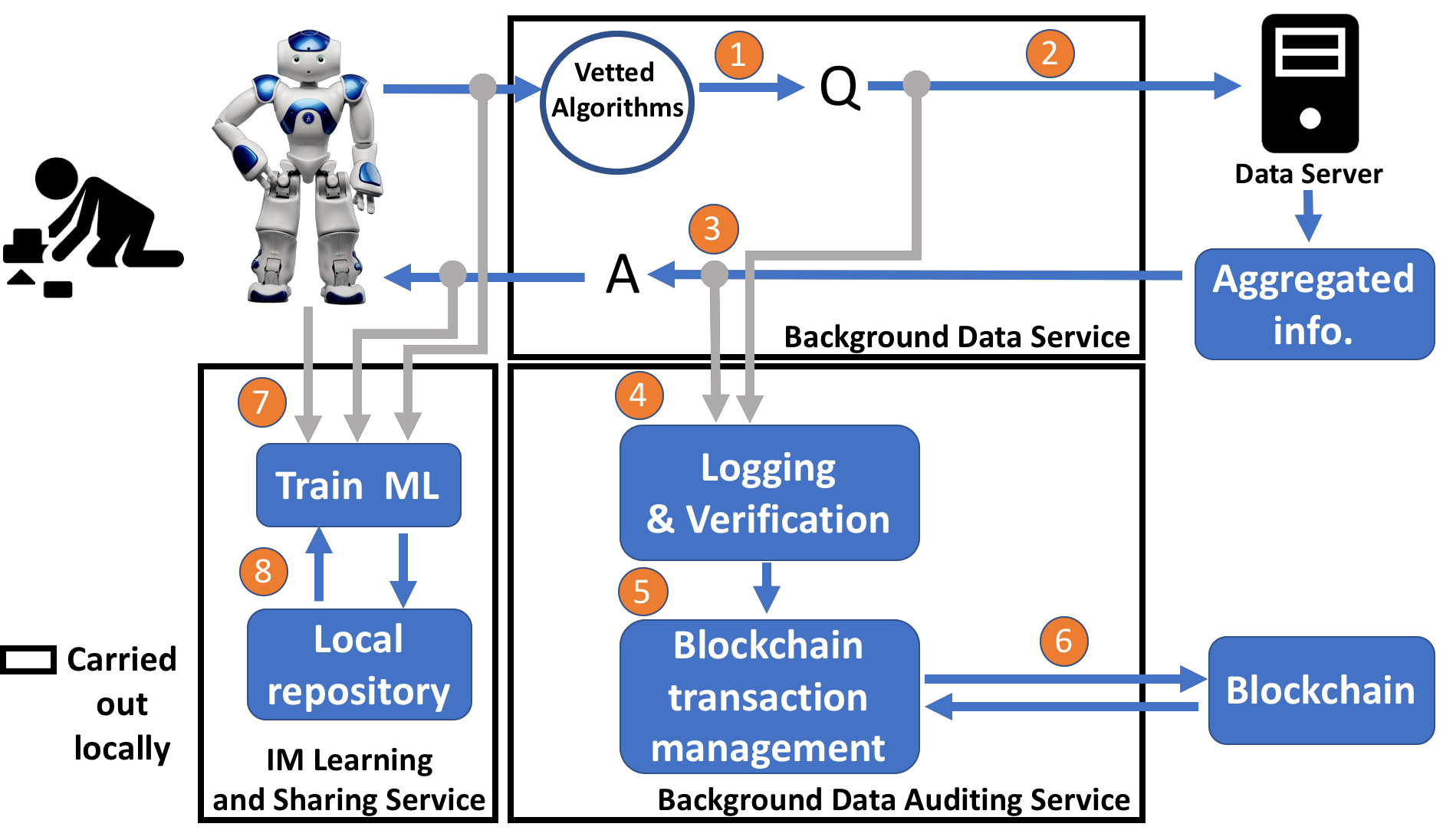}
\caption{System architecture and data flow. The proposed system
  involves three main sections where (A) safe data is retrieved from
  protected databases using the OPAL paradigm, (B) the pairs of
  queries/answers are stored in the blockchain for accountability and
  transparency reasons, (C) the data derived from the therapy is used
  as an input to train and share ML models. All the procedures within
  the black rectangles are carried out in a local fashion (i.e., at
  the robot's own hardware).}
\label{Fig:DataFlow}
\end{figure}

Figure~\ref{Fig:DataFlow} lays out the communication architecture of an
individual robot deployed within a clinical intervention on the target site. This architecture outlines how the robot can gain access to safe BD about the target patient in order to adapt the therapy. It also shows how the access to this information is accounted for, and finally, how the resultant IM can be shared among different robot units in the RoboChain to increase their usability. We break down the proposed data flow into three main components as described below.

\subsection{Background Data Service}
\label{sec:BDService}

\circled{1} - The robot is equipped with a list of secure `vetted'
algorithms to build its queries. The goal here is to ensure
that the algorithms used are free from any kind of bias (e.g., the
culture and gender discrimination, or other types that do not comply
with ethical norms). They should also prevent any unintended side
effects (e.g., the exposure of the patient's behavioral or condition
severity). To accomplish this, the `vetted' algorithms
must be verified beforehand by domain experts. For instance, the robot
may query the general physical condition of the patients undergoing similar clinical interventions in order
to adapt the interactions, which otherwise could adversely
affect the patients. It is also important to note that
this vetting does not guarantee the quality of the output which is a
function of the quality of the input data. As an example, a set of hospitals
that the robot is intending to query about the patient may possess
detailed information about the mental wellbeing of similar patients, but may not return the BD in its original form. Rather, it provides the information about what to be avoided during the intervention
instead of providing the exact details to the robot.


\circled{2} - Queries using one of the `vetted' algorithms are sent to the
data server. These queries may contain information obtained during the
therapy. Specifically, the queries may be designed based on the expert
input (by the caregiver or therapist) and/or by the robot's
sensing of the environment including the patient. For the former, the caregiver administering the intervention at the site may inform the robot about the patient's condition and the type of information needed to
execute/choose the current intervention. For our running example, i.e., the autism therapy, the therapist can decide that the patient needs to work on specific skills (e.g., practicing the eye-gaze exchange). The robot will then try to retrieve relevant
information (through the queries of BD) that can assist him to better adapt
the target IM to this type of social exercise. 

\begin{figure}[htbp]
\centering%
\includegraphics[width=0.5\textwidth]{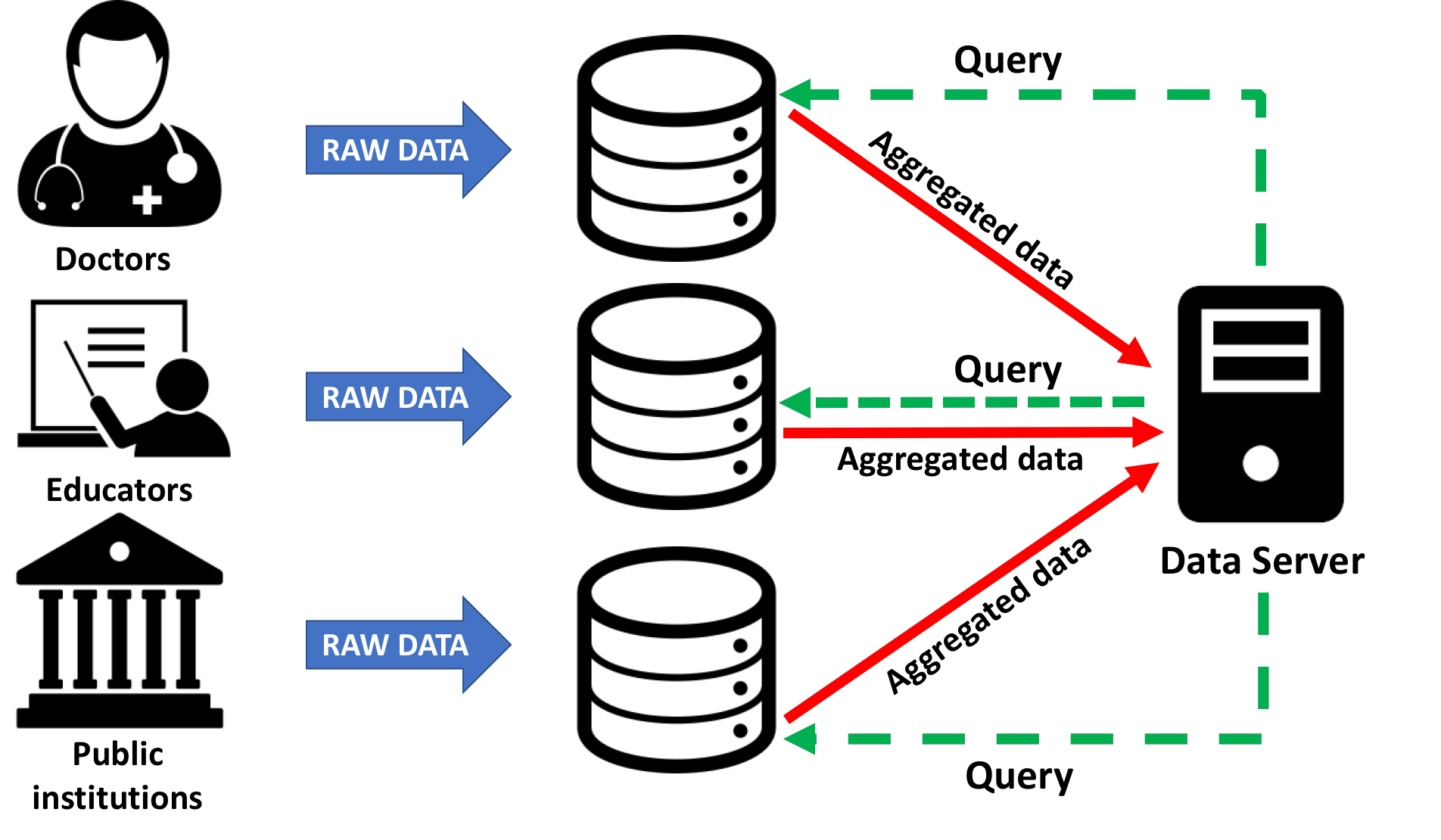}
\caption{Data flow: Trusted parties (e.g., doctors, educators,
  public institutions, etc.) have access to the Background Data (BD) of
  patients. The BD are stored in protected databases that accept
  queries from trusted data servers. Data servers receive the `answers' to
  those queries in the form of aggregated information. Aggregation is
  a useful mechanism to anonymize the information returned.}
\label{Fig:DistributedDatasets}
\end{figure}

To date, the majority of patients' BD
are stored in different datasets scattered across the Internet: belonging to
governments, public institutions, private corporations, etc. Consequently, the most recent advances in network data analysis are starting to address the challenge of creating value from those
datasets without breaching anyone's privacy. OPAL is a framework that proposes a change of paradigm: instead
of copying or sharing raw personal data, algorithms in the form of
queries are sent to the datasets containing the personal information/data. Then,
the queries are executed behind existing firewalls and only anonymous/aggregated information is sent back to the querier. In Figure~\ref{Fig:DistributedDatasets}, a
complete patient profile which may include the patient's medical,
educational, and/or governmental data, is employed to assist the
intervention. Each one of these databases is controlled by a
trusted party, and personal and sensitive data is stored in a `raw'
format from which the BD is retrieved. Then, queries are sent to each
target database and aggregated information is returned (Figure
\ref{Fig:DistributedDatasets}). As mentioned in Section~\ref{bd}, the
aggregated information contains the group-level information about the
patient (e.g., demographics such as culture, age and gender, and
aggregated behavioral assessment scores). This information allows the querier (i.e., a robot) to obtain knowledge about the target patient needed to select the most appropriate
IM/DLMs. The key here is that from the returned information, the patient cannot be identified nor
his/her personal data compromised during the knowledge exchange. In addition,
there is a number of parameters that the robot can passively observe prior to forming the query. For instance, the robot can use computer vision algorithms to automatically infer the patient's age, gender, and the motor abilities.

\circled{3} - An answer with aggregated information is received at the
robot's side. Then, the robot can tailor the IM based on
this. For instance, suppose that the answer is given in
the form: ``For this type of patients, the propensity to negative emotional reactions is $n\%$ higher than in other groups when exercise $x$ is applied.'' By knowing this, the robot can adjust the IM and,
therefore, minimize the risk of exposing the patient to experiencing 
negative emotions.


\subsection{Background Data Auditing Service}

\circled{4} Before the question-answer pair, which is obtained during the
background data service phase, can be passed to the robot, the
background auditing service section of the proposed model captures the
pair to archive it on the blockchain.

\circled{5} Once the question-answer is stored, the system checks how
to include this information in a blockchain transaction. This
requires to encapsulate the information within the right data
structure, depending on the type of blockchain employed. Note that here (a) the blockchain only holds the hash-value, and (b) the complete question-answer is kept by the data service where the raw data is located (as most of today's blockchains cannot encrypt large amounts of data).

\circled{6} The pair of queries-answers used in the system is included in the blockchain. Then, anyone auditing the clinical interventions can inspect the pair of questions-answers used during the patient's therapy. This increases the accountability and transparency of the HRI system, thus, improving its trustworthiness with all parties involved.

\subsection{IM/DLM Learning and Sharing Service}
\label{sec:dshare}
\begin{figure}[htbp]
\centering%
\includegraphics[width=0.45\textwidth]{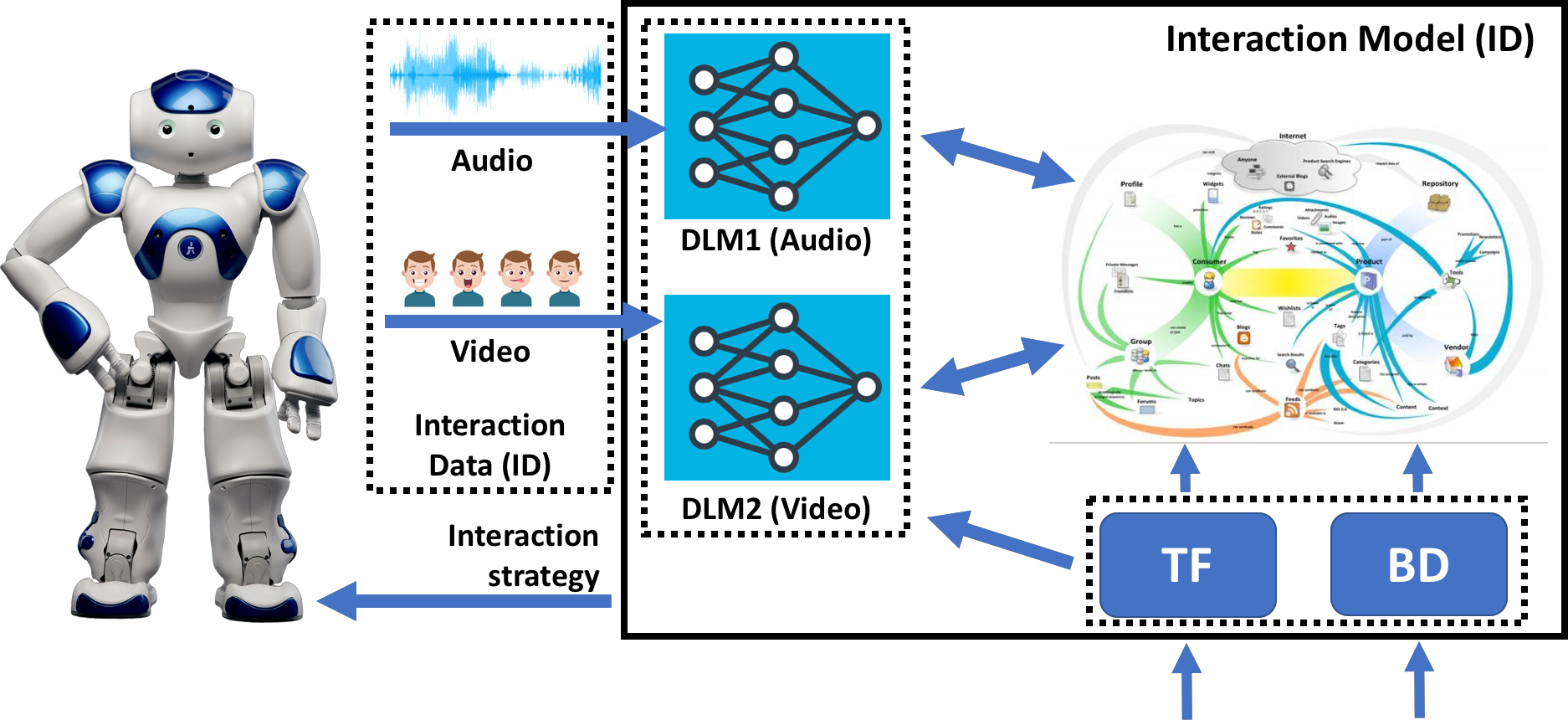}
\caption{Interaction Model (IM). In this example, the robot provides audio-visual recordings of the patient as input to the IM. These are then processed using target DLMs, being part of the existing IM. The outputs of the DLMs (e.g., the patient's stress level and other behavioral cues) are fed into the IM scripts designed for the target intervention. This is further enhanced by the BD and, later, TF, both of which are then used to re-train (update) the DLMs. The output of the IM is the optimal interaction strategy to be performed by the robot as part of an ongoing intervention.}
\label{Fig:DLMS}
\end{figure}

\begin{figure}[htbp]
\centering%
\includegraphics[width=0.5\textwidth]{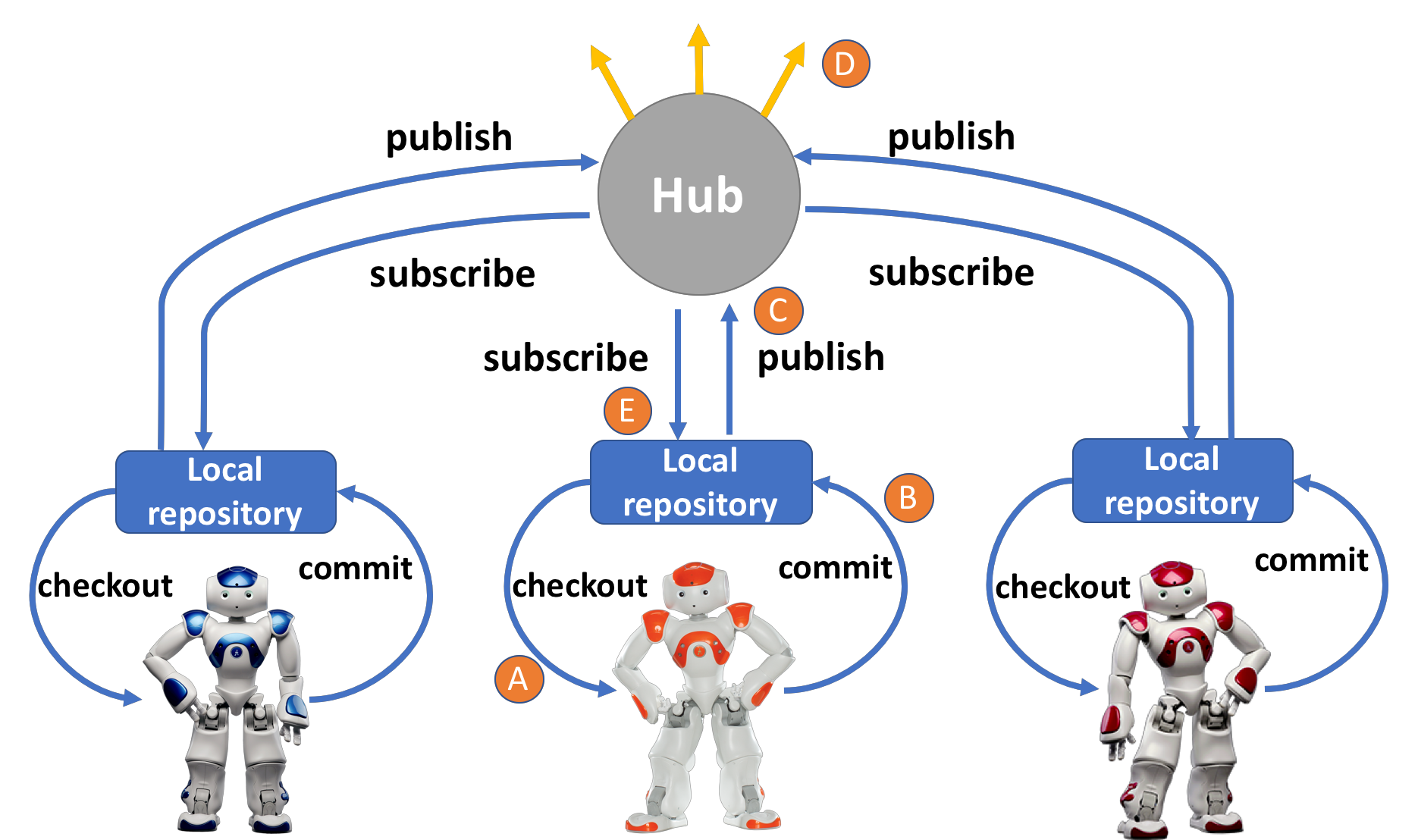}
\caption{A group of robots commit/checkout changes from/to their local
  repository. Robots publish/subscribe to local hubs in order to
  send/get new updates on their DLM. One of the main advantages of
  this approach is that learning always take place locally, while the
  resultant knowledge is distributed globally. Having a local
  repository with the capabilities of a modern source control system
  (e.g., git or mercury) allows robots to calculate differential
  changes on the DLM structure, topology, etc., but also being able to
  roll back to previous versions.}
\label{Fig:LearningTopology}
\end{figure}

\circled{7} As explained in Section~\ref{subsec:InteractionModel},
the robot relies on the DLMs, being part of the target IM, when conducting a clinical intervention. Specifically, the robot checkouts the current DLM from its local hub (Figure~\ref{Fig:LearningTopology} (A)) and uses it to conduct the intervention. During this, the robot stores the information captured by its sensors, producing the ID (e.g., the audio-visual recordings of the patient and his/her responses to the IM). Together with the available BD and TF, these data allow the robot to update/improve the existing DLM models. It is worth to note that updating the IMs is also feasible at this point, however, it requires more input from the domain experts as it concerns the intervention protocols. Specifically, a supervised ML approach is adopted: the ID/BD are used as input to the DLMs, while the TF (e.g., patient's engagement levels) as the target output. Then, the fine-tunning of the DLM parameters to the newly acquired data is accomplished using the standard back-propagation technique and by selecting an optimizer (e.g., Adadelta)~\cite{lecun2015deep}. With these new parameters, the robot is expected to increase its competences and adaptability to target interventions and patients~\cite{Rudovic2017PML}. An example of the data flow to/from IM and its DLMs is depicted in Figure~\ref{Fig:DLMS}. 

\circled{8} The patient data (ID, BD, and TF) collected by robot units at a single site (e.g., a hospital) are being stored at a local hub (Figure~\ref{Fig:LearningTopology}). In this scenario, a hub represents the computing infrastructure of the hospital, care center, etc., where the robot is performing the intervention. After the intervention, the patient data and TF are used together to improve the IMs by re-training their DLMs (as described in \circled{7}). This can be done directly on the hardware of local robot units, and/or on the data-processing servers of the local hub. To this end, different learning strategies can be adopted. For instance, the robots can create, store, and update the target models after each intervention, or after a sufficient amount of data has been collected. The design of specific learning strategies depends on the intervention type, and it is out of the scope of this paper. Then, these new models are committed and stored in the local repository (Figure~\ref{Fig:LearningTopology} (B)) of the local hub. Once created and stored, it is assumed that these new IM/DLMs cannot be used to recreate the raw input data (ID) of the patients. Therefore, all personal data (e.g., images, audio, etc.) remain safe as they do not leave the hub. Furthermore, they are deleted after the models are updated.

The stored IM/DLMs are further deployed locally and assigned a cumulative score based on the TFs derived by validation of this model within new local interventions. As a result, a new Candidate Model (CM) along with its performance score is then created and locked on the local hub. To allow knowledge sharing -- one of the key ingredients of the RoboChain framework -- this new CM is then evaluated by the robot peers operating at other sites, i.e., different hubs within a clinical network. To this end, the local repository (hub) publishes the changes (Figure~\ref{Fig:LearningTopology} (C)) (e.g., the difference between the previous version of the model and the new one). Finally, the hub announces the update to the entire network (Figure~\ref{Fig:LearningTopology} (D)). The goal of this is to assure a fair validation of the CM before it can be adopted as the new version of the IM/DLMs for target intervention.

\begin{figure}[htbp]
\centering%
 \includegraphics[width=0.5\textwidth]{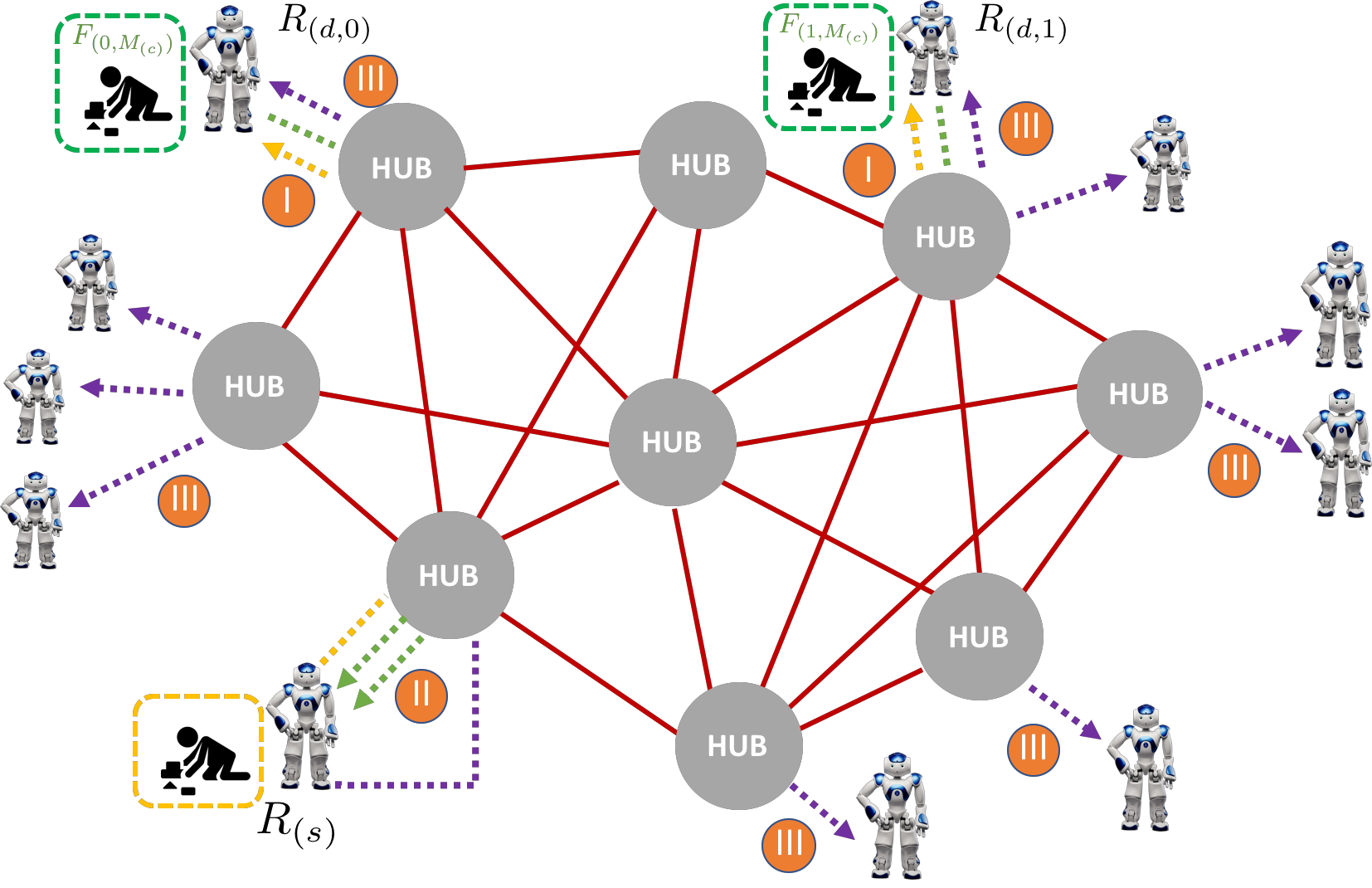}
\caption{A global RoboChain network. Robots connect to local hubs,
  which are interconnected in a sparse network. Robots publish and
  receive notifications whenever new therapy sessions are conducted.
  CMs are proposed by individual robots and validated by
  the network in future therapies. After feedback is provided by
  different peers, a consensus process starts and the utility of the
  CM is evaluated. If the CM outperforms
  previous solutions, the model is acknowledged as the new standard.}
\label{Fig:FederatedNetwork}
\end{figure}

\section{RoboChain and Federated Learning}
The key property of the RoboChain framework is its ability to perform ML operations of the IMs/
DLMs without the need to store the data, acquired during interventions, in a centralized location (e.g., a shared hub within a network of hospitals). This is achieved using the notion of ``Federated Learning'' (FL)~\cite{McMahan2016}, which allows for smarter models, lower latency, and less power consumption, while ensuring the patients' privacy. Furthermore, FL allows the new models to be deployed immediately on the robot units from other sites such as different hospitals in a private network.

Figure~\ref{Fig:FederatedNetwork} depicts the FL approach employed in the RoboChain network. First, as mentioned in Section~\ref{sec:dshare}, the source hub/robot ($R_{(s)})$ `advertises' the new CM by announcing the IM/DLM updates to the entire network. Then,  the destination robots ($R_{(d,i)}$), where $i=1\dots N$ denotes the target sites/hubs, are notified by their local hub that there is an update available in the
network (Figure~\ref{Fig:FederatedNetwork} (I)). This can be achieved by the subscription pipeline they have with their local hub (Figure~\ref{Fig:LearningTopology} (E)). In case the updates are available, the robots can retrieve and apply them to their working
directory.It is worth to note that this does not mean that the robots are committing the received updates to their local repository yet. Once the CM ($M_{(c)}^s$) is adopted from the source hub by the destination robots, the latter will start its evaluation, quantified by the therapists at the destination sites in terms of the TF scores ($F_{(R_{(d,i)},M_{(c)}^s)}$). Additionally, in order to leverage the new local data, the destination hubs can also return the model updates to the source site (obtained in a similar fashion as when creating the CM). These, in turn can be used to construct the new model at the source hub. An example of this approach, but applied in the context of mobile phones, can be found in~\cite{McMahan2016}. 

The next stage in RoboChain is to consolidate the feedback information from the destination hubs/robots (Figure~\ref{Fig:FederatedNetwork} (II)). This can be achieved using time-constraints (i.e., waiting for a pre-defined period of time to receive the feedback), and/or when a target consensus is achieved. For instance, if the feedback score for the CM ($M_{(c)}$) is higher than for the currently accepted model ($M_{(j)}$), i.e., $\overline{F_{(i,M_{(c)}^s)}}>\overline{F_{(i,M_{(j)})}}$. If this is fulfilled, $R_{(s)}$ creates a new model ($M_{(j+1)}$), which is then published to all connected hubs, and committed to their robots' local repositories (Figure~\ref{Fig:FederatedNetwork} (III)). In this way, the new baseline model for future interventions is endorsed by the network. This process can be implemented via modern control version systems (e.g.,
git, mercurial, etc.) in order to store and
share the resulting model configurations (e.g., the DLM topology, hyper-parameters, etc.) obtained after new interventions. This is an important
feature of RoboChain since it allows the robots to rollback to the last consensual version of the model ($M_{(j)}$), in case a consensus did not take place within the network. Moreover, since the robots keep the list of all changes in their local repository, there is a promising research approach in analyzing the metadata available in the updates applied to the repository.

\begin{figure}[htbp]
\centering%
\includegraphics[width=0.5\textwidth]{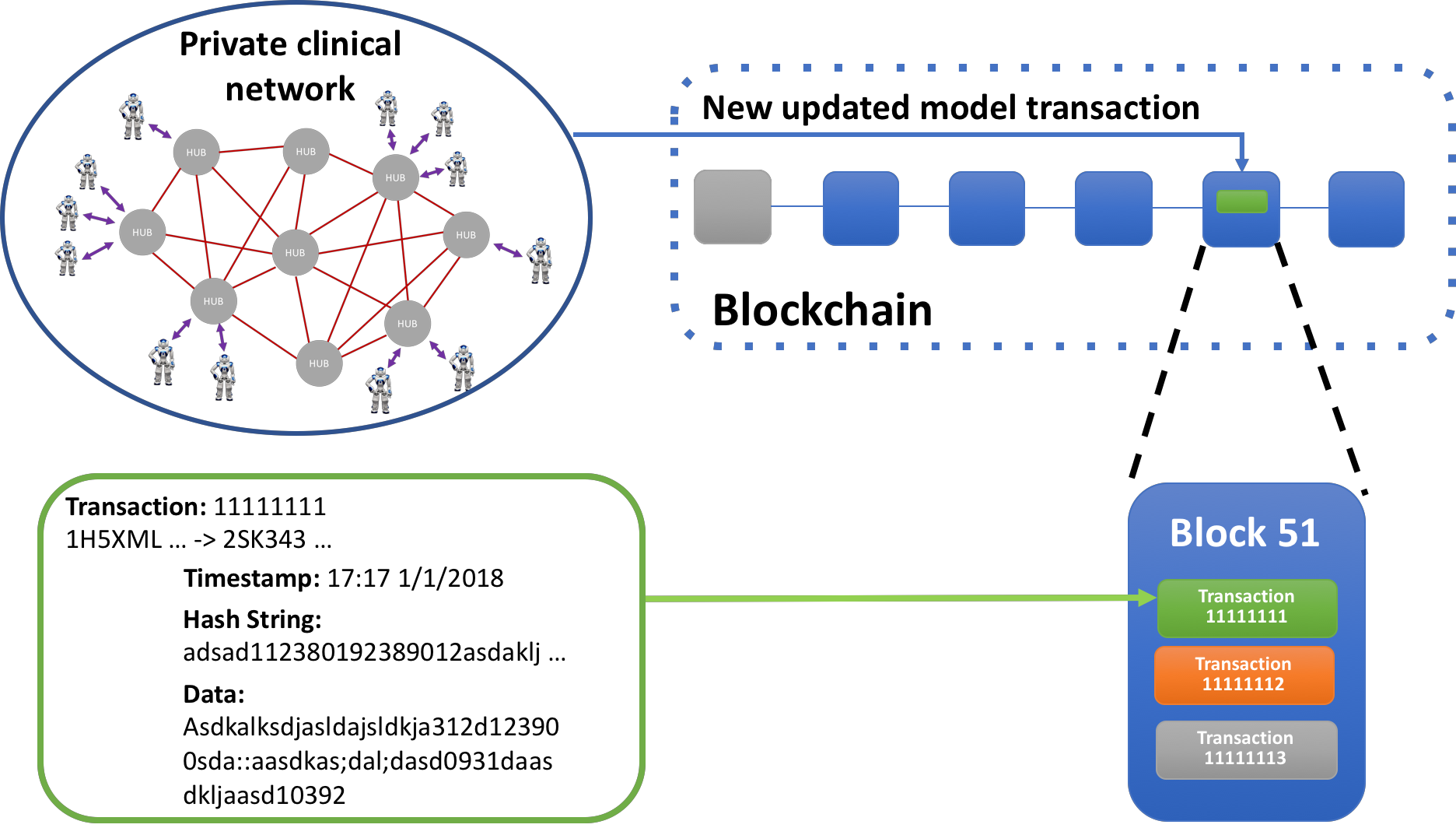}
\caption{Once a $R_{(s)}$ obtains a new candidate model validated by
  the network, $R_{(s)}$ is responsible to send a transaction to a
  blockchain including the key information to verify the consensus
  process. Important information such as timestamps, hash of the new
  model, and encrypted data is provided. Even though this transaction is included in a
  public blockchain, the information within is only readable by the
  participants of the private clinical network.}
\label{Fig:ConsensusResolution}
\end{figure}

In order to notarize and log the creation of the new models and their consensus processes within the network (Figure
\ref{Fig:ConsensusResolution}), $R_{(s)}$ is required to send a
transaction to a blockchain (public, semi-public, or private)
including information such as the timestamp of the global update
broadcast (Figure \ref{Fig:FederatedNetwork} (III)), a hash
string that encapsulates the information about the model update (e.g., differences in the weights, hyper-parameters, etc.), and an encrypted
data field signed by $R_{(s)}$ containing information such as the public IDs of the robots that took part in the consensus process and their correspondent feedback scores. This transaction on the blockchain is necessary to allow participants within the private network (e.g., hospitals, care-centers, etc.) to prove/validate how models were
created, who participated in their consensus process, and when did those transaction take place. It is also important to highlight that the
hash string included in this transaction is useful to check and confirm that the models acquired by robots are indeed the version agreed upon by the network, and not a corrupted version from a
third-party agent. In addition, the encrypted data field signed by $R_{(s)}$ contain sufficient information to allow the users to confirm their participation in the consensus process and check that $R_{(s)}$
was not biased at the moment of promoting $M_{(j)}$. This information is readable by any participant in the private clinical network, since they know the public identifier (i.e., public key) of $R_{(s)}$. By contrast, the peers connected to the blockchain but not members of the private clinical network do not have the means to decrypt this information or identify the nodes involved in the consensus process. This is because all required public/private keys remain
within the boundaries of the clinical private network. Finally, note that the whole consensus reaching process could have been implemented directly on the blockchain via `smart contracts' in order to prevent intruders from `attacking' the network; yet, we assume here that the network access is protected.  

\section{Discussion, Limitations and Future Work}

Until recently, the creation and training of DLMs have been a
computationally-hungry process (e.g., requiring expensive dedicated
hardware such as GPU chips). Because of this, the adoption of these models
in embedded devices (e.g., robots) has been a challenging
task, especially for low-cost robot units. However, not long ago, new
platforms such as TensorFlow
Lite \cite{TFLite2018} have paved the way for deployment of machine learning inference algorithms directly on target devices. This has been made feasible via an improved scheduling process. For instance, the DLM training happens only when
the device is idle, plugged in, and counts with an open connection to
the Internet, so there is no impact on the robot's main tasks. Tools
such as TensorFlow Lite provide a lightweight solution for DLMs
allowing also the federated learning approach, as proposed in this paper. 

The federated learning network
interconnecting the participant parties (e.g., hospitals, care
centers, etc.) and its correspondent robot units is not public. Thus,
it requires a permission of one (e.g., governmental institutions)
or several parties to join and start contributing to the DLM learning
process. We opted for this design approach since we understand that
some form of trust is required in the institutions that deal with a
sensitive task such as autism therapy. For these reasons, we assume
there are no byzantine peers (e.g., malicious partners) within the
network trying to bias or hinder the system operation. However, due to the inclusion of both
transactions in the blockchains, the OPAL query/answer pair and the
announcement of a new DLM adoption, we provide the necessary means to
allow all the participants to check and prove the integrity of the
interaction and learning process in RoboChain. Note also that our current framework assumes that the new models (IM/DLMs) are published by one party at the time; yet, multiple sites can propose new candidate models at the same time. This could easily be  managed by introducing `smart contracts' on the blockchain to attain the synchrony among the sites. Also, we illustrated our framework using a single intervention, but the RoboChain should leverage data from multiple interventions occurring simultaneously at different sites. To tackle this, IMs/DLMs could be structured in a modular form, enhancing each other as more patient data become available. 

One of the aspects left for future work is the adaptation of the proposed system to different ethical frameworks for working with human data~\cite{metcalf2016human}. As mentioned in Section~\ref{sec:BDService}, the use of `vetted' algorithms ensures that the queries created during the therapy are fully aligned with the ethical policies of the relevant regulatory bodies (e.g., Institutional Review Boards (IRB)). This is particularly important when working with healthcare data~\cite{Carsten2016}, as envisioned in RoboChain. It is also important to emphasize that in RoboChain both queries and answers are stored on a blockchain instance, which allows all interested parties to verify the compliance of these ethical norms. However, further research is needed to develop robot controllers that protect the safety (``by default'') of the patients (e.g., due to hardware malfunctioning and/or software bugs). For instance, robot controllers should always ensure that regardless of what type of information/DLMs are obtained through the RoboChain, the physical and mental states of the patients should not adversely be affected in any way.
 
\newpage
To summarize, in this paper, we proposed RoboChain - the first learning framework for tackling the privacy issues in the use of personal data by multiple robots during HRI. We illustrated this framework using autism therapy as an example of an intervention conducted simultaneously by robot units at multiple hospitals. While the RoboChain proposed here offers the main principles for secure data and models sharing between multiple robot units and their sites, in future we plan to empirically evaluate this concept. To this end, in the first stage we aim to collect the intervention data from several hospitals in order to test the RoboChain framework in a simulated data-exchange scenario. In the second stage, we aim to have the RoboChain system running in real time on a private network. One of the main challenges that we envision in this process is how to enable efficient and real-time learning of IMs/DLMs. We expect that for this, existing ML approaches will need to be adapted so that they can efficiently communicate/be integrated with the OPAL and blockchain technologies.  Also, how to extend the RoboChain framework so that it can simultaneously handle multiple private networks, in order to further increase the learning efficacy or interface with mobile health devices, are other promising directions to pursue. 


\section*{Acknowledgment}

This project has received funding from the European Union’s Horizon
2020 research and innovation programme under the Marie
Skłodowska-Curie grant agreement No. 751615 and No. 701236.

\bibliographystyle{IEEEtran}
\bibliography{References}
\end{document}